\documentclass[final]{cvpr}
\usepackage{times}
\usepackage{epsfig}
\usepackage{graphicx}
\usepackage{amsmath}
\usepackage{amssymb}
\usepackage{array}
\usepackage[table]{xcolor}
\usepackage{color}
\usepackage{colortbl}

\newcommand{\modelname}{Affect2MM}
\usepackage{multirow, makecell}
\usepackage{subcaption}
\usepackage{booktabs}
\usepackage{amsmath}
\usepackage{amssymb}
\usepackage{amsfonts}
\usepackage{enumitem}
\usepackage{bbm}
\usepackage{soul}
\usepackage{dsfont}
\usepackage{bbold} % all ten digits with \mathbb
\usepackage{hhline}

\DeclareMathOperator*{\argmin}{arg\,min}
\newcommand{\shorteq}{%
  \settowidth{\@tempdima}{-}% Width of hyphen
  \resizebox{\@tempdima}{\height}{=}%
}

\newcolumntype{L}[1]{>{\raggedright\let\newline\\\arraybackslash\hspace{0pt}}m{#1}}
\newcolumntype{C}[1]{>{\centering\let\newline\\\arraybackslash\hspace{0pt}}m{#1}}
\newcolumntype{R}[1]{>{\raggedleft\let\newline\\\arraybackslash\hspace{0pt}}m{#1}}

\usepackage{amsthm}

\newtheorem{problem}{Problem}[section]

\usepackage[pagebackref=true,breaklinks=true,colorlinks=true, citecolor=magenta, bookmarks=false]{hyperref}

\def\cvprPaperID{11090} % *** Enter the CVPR Paper ID here
\def\confYear{CVPR 2021}

\begin{document}
%%%%%%%%% TITLE
\title{Affect2MM: Affective Analysis of Multimedia Content Using Emotion Causality}
\author{Trisha Mittal\textsuperscript{\rm 1}, Puneet Mathur\textsuperscript{\rm 1}, {Aniket Bera\textsuperscript{\rm 1}, Dinesh Manocha\textsuperscript{\rm 1,\rm 2}}\\
\textsuperscript{\rm 1}Department of Computer Science, University of Maryland, College Park, USA\\ 
\textsuperscript{\rm 2}Department of Electrical and Computer Engineering, University of Maryland, College Park, USA\\ 
\{trisha, puneetm, bera, dmanocha\}@umd.edu\\ 
Project URL: \url{https://gamma.umd.edu/affect2mm}
}

\maketitle
%\thispagestyle{empty}

%%%%%%%%% ABSTRACT %%%%%%%%%
\begin{abstract}
We present \modelname, a learning method for time-series emotion prediction for multimedia content. Our goal is to automatically capture the varying emotions depicted by characters in real-life human-centric situations and behaviors. We use the ideas from emotion causation theories to computationally model and determine the emotional state evoked in clips of movies. \modelname~explicitly models the temporal causality using attention-based methods and Granger causality. We use a variety of components like facial features of actors involved, scene understanding, visual aesthetics, action/situation description, and movie script to obtain an affective-rich representation to understand and perceive the scene. We use an LSTM-based learning model for emotion perception. To evaluate our method, we analyze and compare our performance on three datasets, SENDv1, MovieGraphs, and the LIRIS-ACCEDE dataset, and observe an average of $10-15$\% increase in the performance over SOTA methods for all three datasets. 
\end{abstract}

%%%%%%%%% INTRODUCTION %%%%%%%%%
\section{Introduction}
\label{sec:introduction}
In affective computing, perceiving the emotions conveyed in images and videos has found applications in digital content management~\cite{joshi2014aesthetics, wang2015sentiment}, digital marketing~\cite{mcduff2014predicting, Hussain2017AutomaticUO, ye2018advise}, education~\cite{downs2008effectiveness, alqahtani2019comparison}, and healthcare~\cite{cohn2009detecting}. Such applications have resulted in automated ranking systems, indexing systems~\cite{wang2006affective, wiley2003emotion, 10.3389/fpsyg.2019.01935}, and more personalized movie recommendation systems~\cite{oliveira2011ifelt}.   
% With a continuously growing audience for social media and digital platforms, commercial content creators and advertisement agencies often indulge in media measurement to evaluate their content~\cite{mcduff2013predicting,mcduff2014predicting, mcduff2016applications}. Incorporating target customer behavior~(affect, emotion, and other cues) in testing media content has helped connect to the audience effectively~\cite{vedula2017multimodal}. Along similar lines,  affective analysis of movies has been a problem of interest in the community~\cite{quan2018frame, jin2017thuhcsi}. --- RELATED WORK
% The goal here is to understand the emotions that the movies invoke in the audience.
Affective analysis of movies has been a problem of interest in the community~\cite{quan2018frame, jin2017thuhcsi}, along similar lines. In our work, we explore the problem of affective analysis of movies with the goal of understanding the emotions that the movies invoke in the audience. 
% Such an analysis can help in an automated ranking, indexing~\cite{wang2006affective, wiley2003emotion, 10.3389/fpsyg.2019.01935} and more personalized movie recommendation systems~\cite{oliveira2011ifelt}.   
%%%%%%%%%%%%%%%%%%%%%%%%%%%%
\begin{figure}[t]
    \centering
    \includegraphics[width = \columnwidth]{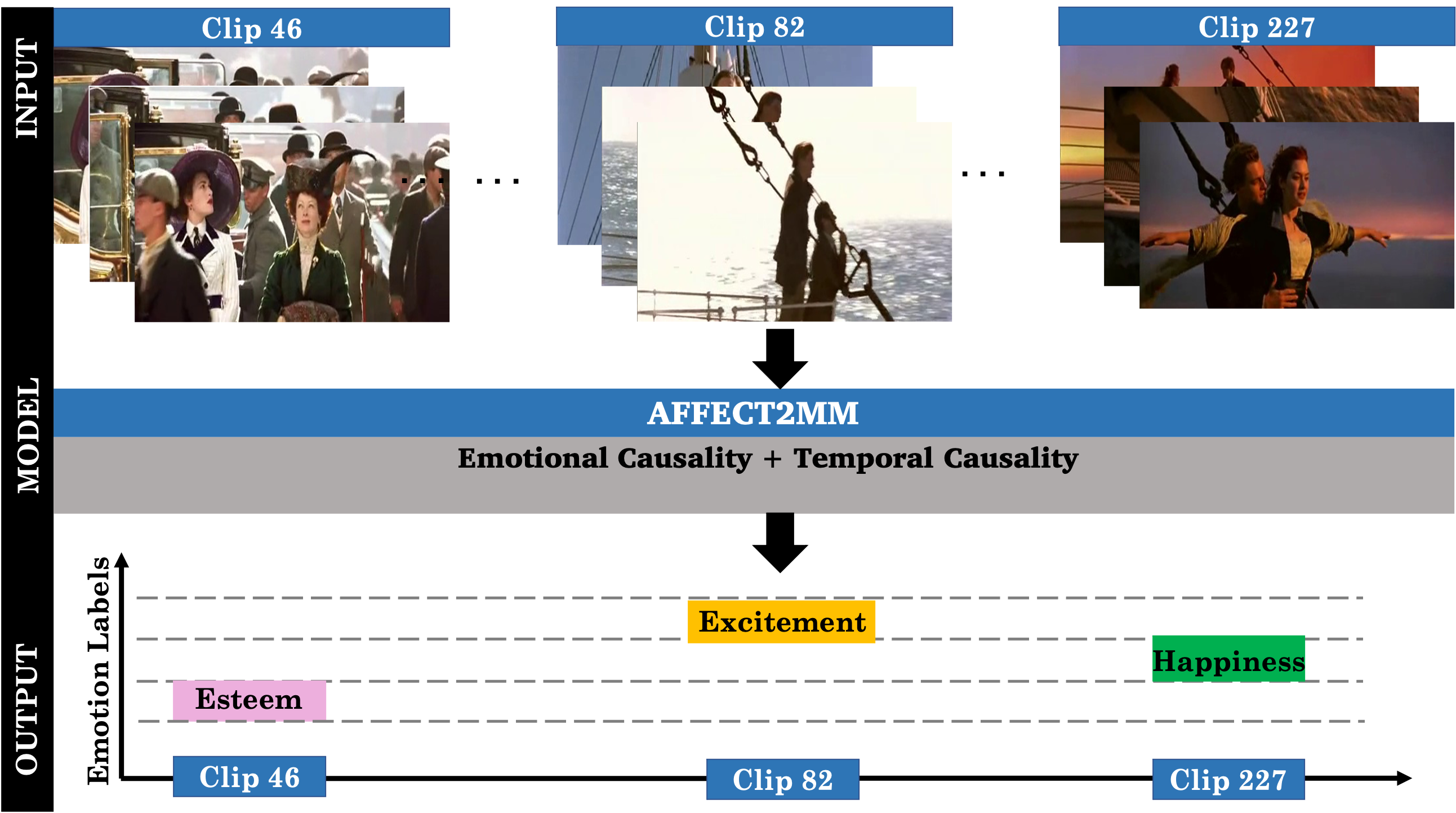}
    \caption{\small{\textbf{Time-Series Emotion Perception Model: }We present \modelname, a learning model for time-series emotion perception for movies. We input a multimedia content in the form of multiple clips and predict emotion labels for each clip. \modelname~is based on the theory of emotion causation and also borrows idea for temporal causality. We show some example clip-frames from the movie `Titanic', a part of the MovieGraphs Dataset and corresponding emotion labels.}}
    \label{fig:cover}
    \vspace{-15pt}
\end{figure}
%%%%%%%%%%%%%%%%%%%%%%%%%%%%%%

There has been a growing interest~\cite{ong2019modeling} in dynamically modeling the emotions over time~(`time series emotion recognition' among the affective computing community. This underlying problem uses temporally continuous data~(facial features, speech feature, or other modality features) from multimedia content as input and predicts the emotion labels at multiple timestamps~(clips) of the input. To aid in solving this time-series problem, several time-series emotion datasets have been proposed~\cite{mckeown2010semaine,Hussain2017AutomaticUO,trigeorgis2016adieu, kossaifi2019sewa,barros2018omg,ong2019modeling}. While these datasets focus more on single-person emotional narratives recorded in controlled settings, multimedia datasets~(movie databases) like LIRIS-ACCEDE~\cite{baveye2015liris} and MovieGraphs~\cite{moviegraphs}~(annotated for per-clip emotion labels) are also being explored for time-series emotion perception tasks. 

% ~\cite{baveye2015liris, moviegraphs}. Movies, as a time-series multimedia content, model multiple human-centric situations and are temporally very long, but coherent sequences. To be able to reason about emotions invoked at various clips of the movie, it is important to develop a causal understanding of the story. Current approaches for time-series emotion prediction, however, do not currently model causality in time-series multimedia content.
% To this end, datasets like SEMAINE~\cite{mckeown2010semaine}, AViD-Corpus~\cite{Hussain2017AutomaticUO}, RECOLA~\cite{trigeorgis2016adieu}, SEWA~\cite{kossaifi2019sewa}, OMG-Emotion~\cite{barros2018omg} and SEND~\cite{ong2019modeling} have been commonly evaluated on. While these datasets focus more on single-person emotional narratives recorded in controlled settings, multimedia datasets~(movie databases) like LIRIS-ACCEDE~\cite{baveye2015liris} and MovieGraphs~\cite{moviegraphs}~(annotated for per-clip emotion labels) are also being explored for time-series emotion perception tasks.

There have been various efforts to understand how humans reason and interpret emotions resulting in various theories of emotion causation based on physiological, neurological, and cognitive frameworks. One such theory is the ``emotional causality''~\cite{films-causal} that has been developed from the Causal Theory of Perception~\cite{hyman1992causal} and Conceptual Metaphor Theory~\cite{athanasiadou2010speaking, niemeier1997language, kovecses2003metaphor}. ``Emotional Causality'' refers to the understanding that an experience of emotion is embedded in a chain of events comprising of an (a) outer event; (b) an emotional state; and (c) a physiological response. Few works have explored such emotional causality for emotion perception in multimedia tasks.

Movies, as a time-series multimedia content, model multiple human-centric situations and are temporally very long, but coherent sequences. To be able to reason about emotions invoked at various clips of the movie, it is important to develop a causal understanding of the story. Generic methods of handling such temporality include recurrent neural networks~\cite{landis1977measurement, soleymani2014continuous}, attention-mechanisms~\cite{chen2014deepsentibank}, graph modeling~\cite{graph-1}, and statistical methods like Granger causality~\cite{gc}. Explicit modeling of causality in the context of time-series emotion perception has been relatively unexplored. 
Emotion labels have been explored extensively, both as discrete~\cite{Kosti_2017_CVPR_Workshops} and continuous~\cite{vad}, in affective analysis. The Valence-Arousal-Dominance (VAD) model~\cite{vad} is used for representing emotions in a continuous space on a $3$D plane with independent axes for valence, arousal, and dominance values. The Valence axis indicates how pleasant (vs. unpleasant) the emotion is; the Arousal axis indicates how high (or low) the physiological intensity of the emotion is, and the dominance axis indicates how much the emotion is tied to the assertion of high (vs. low) social status. A combination of $3$ values picked from each axis represents a categorical emotion like `angry' or `sad', much like how an $(x,y,z)$ point represents a physical location in $3$-D Euclidean space. Various transformations~\cite{discrete_to_continuous_1,discrete_to_continuous_2} can be used to map discrete emotion labels to the VAD space. In this work, we work with both continuous emotion labels and discrete emotion labels. 
% While valence~(pleasant (vs. unpleasant) the emotion is), arousal~(high (vs. low) physiological intensity), and dominance~(assertion of high (vs. low) social status) indicate different aspects, they together indicate an observed emotion.  
% In fact, discrete emotions can actually be mapped back to the VAD space through various transformations~\cite{discrete_to_continuous_1,discrete_to_continuous_2}. In this work, we work with both continuous emotion labels and discrete emotion labels. 
 
\textbf{Main Contributions:} The following are the novel contributions of our work.
\begin{enumerate}[noitemsep]
    \item We present~\modelname, a learning-based method for capturing the dynamics of emotion over time. \modelname~aligns with the psychological theory of ``emotional causality'' to better model the emotions evoked by each clip of a movie. 
    \item To better model the temporal causality in movies for long-range multimedia content like movies, we use attention methods and Granger causality to explicitly model the temporal causality~(between clips in movies). Our approach can be used for predicting both continuous emotion labels~(valence and arousal) and also discrete class labels.  
\end{enumerate}

We evaluate our method on two movie datasets, MovieGraphs~\cite{moviegraphs} and the LIRIS-ACCEDE~\cite{baveye2015liris} dataset. To showcase our method's generalizability, we also evaluate and compare our method on the SENDv1~\cite{ong2019modeling} dataset, a single-person emotional narratives dataset. 

%%%%%%%%% RELATED WORK %%%%%%%%%
\section{Related Work}
\label{sec:relatedwork}
In this section, we summarize prior work done in related domains. We first look into available literature in affective analysis of multimedia content and various applications in Section~\ref{subsec:digital-content}. In Section~\ref{subsec:visual-affective}, we discuss the visual affective representations that have previously been explored for related tasks.  We also discuss conventional methods to model temporality in Section~\ref{subsec:temporal-causal}. In Section~\ref{subsec:psych}, we discuss ``emotional causality'' and other theories of emotion, as suggested in Psychology literature and the need to align computation models with these theories. 
%%%%%%%%%%%%%%%%%%%%%%%%%%%%%%%%%%%%%%%%%%%%%%%%%%%%
\subsection{Affective Analysis of Multimedia Content}
\label{subsec:digital-content}
Various approaches have explored understanding emotions evoked by multimedia content. Chen et al.~\cite{chen2014deepsentibank}, Ali et al.~\cite{ali2017high}, Wei et al.~\cite{wei2020learning} have performed affective multi-class classification on images collected from popular websites. Pilli et al.~\cite{pilli2020predicting}, Hussain et al.~\cite{hussain2017automatic}, and Zhang et al.~\cite{ zhang2020look} studied predicting sentiments in image advertisements. Vedula et al.~\cite{vedula2017multimodal} extended this idea and developed an advertisement recommendation system using sentiments in advertisement content.
Understanding the relationship between emotional responses to content has been learned by recording viewer responses in many controlled use studies. Based on such studies, researchers have used facial responses~\cite{mcduff2014predicting, kassam2010assessment, micu2010measurable}, facial electromyography~(EMG)~\cite{micu2010measurable}, electroencephalogram~(EEG), pupillary response and gaze~\cite{teixeira2014and, soleymani2011multimodal}, smile~\cite{mcduff2013predicting, teixeira2014and, yang2014zapping}. Similarly, Philippot~\cite{philippot1993inducing} and Gross and Levenson~\cite{gross1995emotion} were the first ones to propose a small dataset of clips from films with participants responses to them in controlled lab settings. McDuff et al.~\cite{mcduff2014predicting} have recognized the constraint of collecting such data in controlled settings and have proposed collecting large-scale viewer data using webcams. Other movie-based datasets with some affective annotations that were used are HUMAINE~\cite{humaine}, FilmStim~\cite{filmstim}, DEAP~\cite{deap}, MAHNOB-HCI~\cite{soleymani2011multimodal}, EMDB~\cite{emdb}, MovieGraphs~\cite{moviegraphs} and LIRIS-ACCEDE~\cite{baveye2015liris}. In our work, we evaluate our method on some of these datasets.
%%%%%%%%%%%%%%%%%%%%%%%%%%%%%%%%%%%%%%%%%%%%%%
\subsection{Visual Affective Rich Representation}
\label{subsec:visual-affective}
Affective analysis of multimedia from the viewer's response~(face and body posture reactions, EEG, and ECG signals) is not scalable due to lack of data. Subsequent efforts are being made to perform the same analysis using cues directly from the content: images/videos/movies. Wei et al.~\cite{wei2020learning}, and Panda et al.~\cite{panda2018contemplating} report that general visual feature extractions used for standard vision tasks~(object recognition) do not scale up in terms of performance for emotion-related tasks. Scene and place descriptors extracted from the image/frames have been explored~\cite{ali2017high} to understand the affective component better. Researchers~\cite{quan2018frame,jin2017thuhcsi,batziou2018visual} have also focused on the visual aesthetics of the multimedia content to understand how they affect the evocation of emotions from viewers. Zhao et al.~\cite{zhao2019video} have also used background music to analyze the affective state. In our work, we present a total of $6$ features that we believe can help in a better understanding of multimedia content. 
%%%%%%%%%%%%%%%%%%%%%%%%%%%%%%%%%%%%%%%%%%%%%%%%%%%%
%%%%%%%%%%%%%%%%%%%%%%%%%%%%%%%%%%%%%%%%%%%%%%
\subsection{Modeling Temporality in Time-Series Models}
\label{subsec:temporal-causal}
While emotional causality guides us to study the emotional state of a clip, it becomes important to keep track of and model the temporality in emotions in long multimedia content like movies. While recurrent neural network architectures are inherently designed to keep track of such dependencies, explicitly modeling temporal causality has been a norm. Attention-based methods and their many variants~\cite{cheng2016long,chen2014deepsentibank} have been used before to help models learn important parts to ``attend to'' in time series data. Furthermore, most commonly, causality is frequently studied using graph structures and modeling~\cite{graph-1,graph-2}. More conventionally, statistical methods like Granger causality~\cite{gc} have been used to quantify the dependence of past events into the future time series. More recently~\cite{gc2, tank2018neural}, there have been multiple attempts to incorporate similar behaviors in neural networks. 
%%%%%%%%%%%%%%%%%%%%%%%%%%%%%%%%%%%%%%%
\subsection{Theory of Emotions: A Look Into Psychology}
\label{subsec:psych}
Some of the major theories of emotion that reason causation of emotion can be grouped into physiological, neurological, and cognitive theories. Physiological theories like the James-Lange Theory~\cite{10.1093/mind/os-IX.34.188} and the Cannon-Bard Theory~\cite{cannon1927james} suggest an external stimulus leads to a physiological reaction, and the emotional reaction is dependent on how the physical reaction is interpreted. Schachter-Singer Theory~\cite{schachter1962cognitive} is based on cognition propose, a two-factor theory according to which a stimulus leads to a physiological response that is then cognitively interpreted and labeled, resulting in an emotion. Many more such theories attempt to understand how humans think about emotional states, also called ``affective cognition'' by ~\cite{ong2019computational}. They also provide a taxonomy of inferences within affective cognition and also provide a model of the intuitive theory of emotion modeled as a Bayesian network. Another term scholars have used to understand this domain is ``emotional causality''~\cite{films-causal}.  
%%%%%%%%%%%%%%%%%%%%%%%%%%%%%%%%%%%%%%%

%%%%%%%%% APPROACH %%%%%%%%%
\section{Background and Overview}
\label{sec:background}
We formally state our problem in Section~\ref{subsec:problem-setup}. We then give a brief overview of co-attention mechanism~(Section~\ref{subsec:co-attention}) and Granger causality~(Section~\ref{subsec:gc}).

%%%%%%%%%%%%%%%%%%%%%%%%%%%%%%%%%%%%%%%%%
%%%%%%%%%%%%%%%%%%%%%%%%%%%%%%%%%%%%%%%%%
\begin{figure*}[t]
\centering
\includegraphics[width=.8\linewidth]{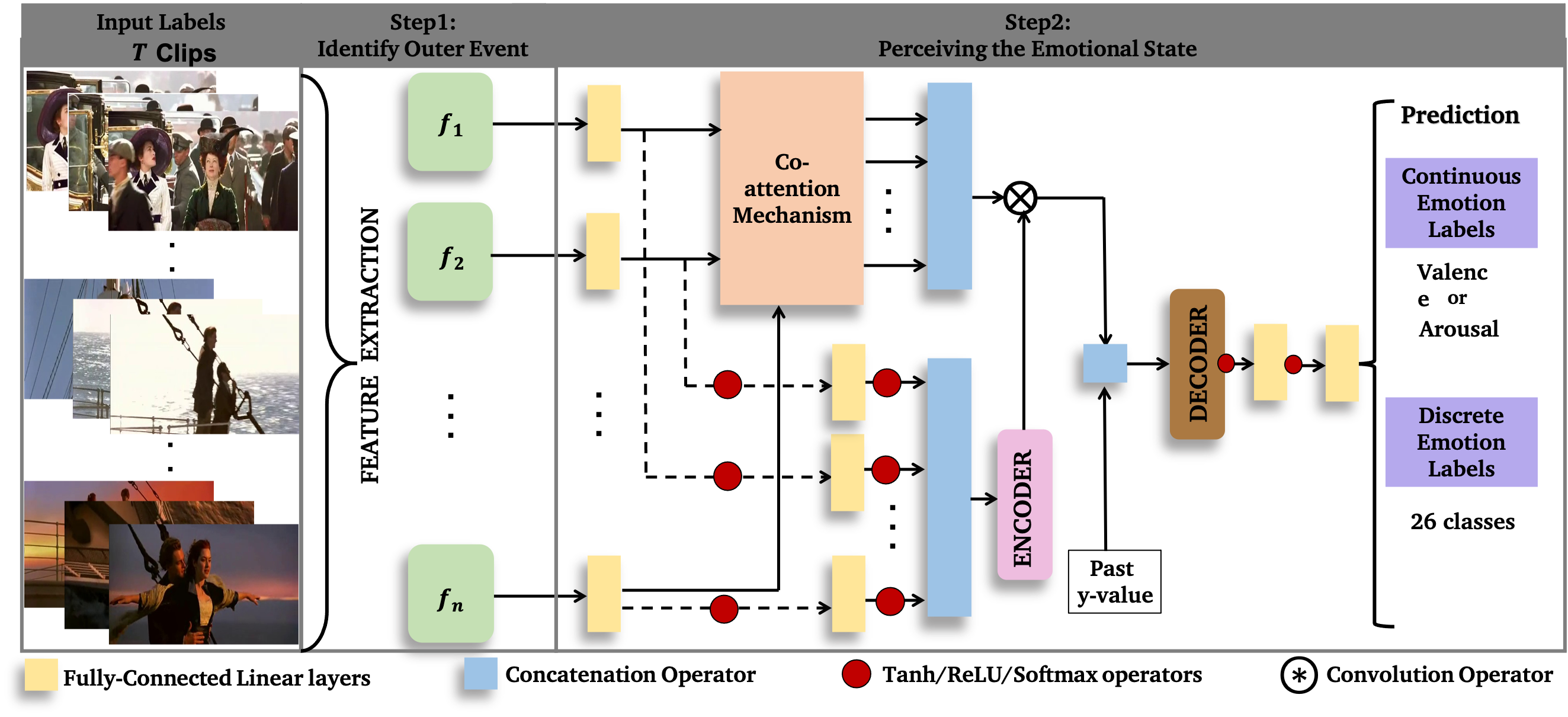}
\caption{\small{\textbf{\modelname: } We use the components of Emotional Causality to infer the emotional state depicted in the multimedia content. Given the input $T$ clips, we first extract features, $f_1 \dots f_n$, to build the affective-rich representation and help identify the outer event/cause. We then pass these feature representations through a co-attention mechanism, cLSTM-based encoder, and decoder to perceive the emotional state in either continuous values~(valence/arousal) or discrete labels. The networks consist of fully-connected layers~(yellow), concat layers~(blue), and non-linearity~(red dots). }}
\label{fig:network}
\vspace{-15pt}
\end{figure*}

%%%%%%%%%%%%%%%%%%%%%%%%%%%%%%%%%%%%%%%%%
\subsection{Problem Statement}
\label{subsec:problem-setup}

We consider multimedia content, which can be any image, video, or audio in the form of multiple \textit{clips}. Each clip, $\mathcal{C}$, is a short sequence of frames that contains information arising from multiple modalities such as facial expressions of the actors, their speech, the transcribed dialogues, the visual aesthetics of the scene, the description of the scene, and so on. Our goal is to predict the emotion label of each clip using the information contained in the corresponding frames. More formally,

\begin{problem}
Given a set of clips spanning $T$ time-steps, $\mathcal{C}^{1:T} = \{ f_1^{1:T},f_2^{1:T}, \ldots f_p^{1:T}\}$, where $f_i^{1:T}$ denotes the $i^\textrm{th}$ feature for each clip, we predict the emotion label, $y$, for each clip, denoted by $y^{1:T}$.
\end{problem}

\noindent Our formulation allows $y$ to be general in that it may represent either categorical emotion labels (``happy'', ``sad'', $\ldots$) or continuous real-valued labels, depending on the dataset. In this work, $y$ can be either represent $26$ categorical labels or one of the $2$ real-valued labels - valence and arousal.

% Given an input multimedia content, which can be an image, video, or audio represented by $X$, the goal is to predict the corresponding categorical labels of emotions or real-valued scores, $Y$. More formally stated, given $n$ pairs of training sequences $\left \{ \left ( X^{k}_{1:T_k}, Y^{k}_{1:T_k},\right ), 1 \leq k \leq n \right \}$ where $T_k$ is the final time point of sequence $k$, the goal is to train such a model that can predict the sequence of outputs $Y^{j}_{1:T_j}$ given a new input sequence $X^{j}_{1:T_j}$, for some $j > n$

%%%%%%%%%%%%%%%%%%%%%%%%%%%%%%%%%%%%%%%%%
\subsection{Co-attention Mechanism}
\label{subsec:co-attention}
Attention mechanisms~\cite{bahdanau2016neural} in neural networks have a long history in both the NLP and vision communities. The broad idea is to allow the learning model to only attend to relevant parts of the input rather than encoding the entire input. While intuitive, modeling it requires the computation of attention scores, a weighted matrix depicting the dependence of all outputs on each of the inputs. These attention scores are then used for making inferences. There have been many variants like self-attention~\cite{chen2014deepsentibank}, hierarchical and nested attention~\cite{yang-etal-2016-hierarchical}, and attention flow~\cite{seo2018bidirectional}. One such variant is the co-attention mechanism~\cite{xiong2018dynamic}.  

The co-attention mechanism calculates the shallow semantic similarity between the two inputs and uses that as a reference. Given two inputs $(\big\{u_{p}^{t}\big\}_{t=1}^{t=T}, \big\{u_{q}^{t}\big\}_{t=1}^{t=T})$, where $T$ is the timestep, co-attention aligns them by constructing a soft-alignment matrix $S$. Each $(i,j)$ entry of the matrix $S$ is the multiplication of the $\tanh$ activation for both the inputs.
$$S_{i,j} = \tanh(w_{p}u_{p}^{i}) \cdot \tanh(w_{q}u_{q}^{j})$$

We feed these inputs through a single-layer neural network followed by a softmax function to generate the attention distribution $\alpha$~\cite{Zhang2018AdaptiveCN}. 
\begin{align}
\begin{split}
z & = \tanh \left ( w_{p}u_{p}^{i} \oplus w_{q}u_{q}^{j}\right )\\
\alpha &= \text{softmax} \left( w_{\alpha}z \right)
\end{split}
\label{eq:co-attention-mechanism}
\end{align}
We use $\oplus$ to denote the concatenation of the two features. Based on the attention distribution, the attended/relevant parts can be obtained as follows: 
$$\hat{u}_{q}^{j} = \sum_{i}\alpha_{i,j} \cdot u_{p}^{i}$$

%%%%%%%%%%%%%%%%%%%%%%%%%%%%%%%%%%%%%%%%%
\subsection{Granger Causality~(GC)}
\label{subsec:gc}
Granger causality~\cite{gc} is a way of quantifying the extent to which the past activity of some time-series data is predictive of an other time-series data. In our case, we use the term ``time-series data'' to refer to an element of $\mathcal{C}^{1:T} = \{ f_1^{1:T},f_2^{1:T}, \ldots f_p^{1:T}  \}$. The $p$ features may include facial action units of the actors involved, the audio features, embeddings of the text/transcript, scene descriptions, action or situation descriptions, visual aesthetics of the clips, etc. In other words, the sequence of, say, facial features across $T$ clips, $f_1^{1:T}$ is a time-series modality data. We use GC to reason the causality between different modalities connected across time. If there exists a causality between $f_1^{1:T}$ and $f_2^{1:T}$, then we say that $f_1$ \textit{Granger-causes} $f_2$. 

In our approach, we first explore the existence of Granger causality between time-series modalities in temporal clips. If GC is found to exist between two modalities, then we show that it can be used to improve the accuracy of any general emotion recognition system.
\paragraph{Existence of GC in LSTMs:} The most widely-used model to estimate GC in linear systems is the Vector AutoRegressive (VAR) model~\cite{var1, var2}. However, solutions for the linear models do not scale to non-linear time series data~\cite{Tersvirta2011ModellingNE, gc2}. Tank et al.~\cite{tank2018neural} address non-linear GC in neural network architectures like Multi-Layer Perceptron~(MLP) and LSTMs by introducing a new LSTM architecture called component-wise LSTM~(cLSTM), which models each time-series modality through independent LSTM networks. More formally, consider input
\begin{equation}
X_{1:T} = \Pi \left( \begin{bmatrix} 
    f_{1,1} & f_{1,2} & \dots & f_{1,T} \\
    f_{2,1} & f_{2,2} & \dots & f_{2,T} \\
    \vdots &\vdots &\ddots & \vdots\\
    f_{p,1} & f_{p,2} & \dots & f_{p,T} \\
    \end{bmatrix} \right )
    \label{eq: GC_representation}
\end{equation}
\noindent where the $j^\textrm{th}$ row of $X_{1:T}$ is denoted as $X^{(j)}$ and linear transformation $\Pi$ is applied row-wise (more details in Section~\ref{subsec:encoder}). To  check the existence of GC, we begin by passing $X_{1:T}$ to the cLSTM which is made up of $p$ separate LSTMs, $L^{(j)}$, $j=1,2,\ldots, p$, each operating on the input $X_{1:T}$. In our system, this stage is executed using Equation~\ref{eq: encoder}. Then, following the theoretical framework put forth in Tank et al.~\cite{tank2018neural}, we check for GC by solving the following optimization problem using line search \cite{armijo1966}: 
\begin{equation}
    W^{*(j)} = \argmin_{W^{(j)}} \sum_t (x_{i,t} - L^{(j)}(X_{1:T}))^2 + \lambda\sum_{k=1}^p \left\Vert W^{(j)}_{:k}\right\Vert_2,
    \label{eq: GC_equation}
\end{equation}
\noindent If $\left\Vert W^{*(j)}_{:k}\right\Vert_2 = 0$, then the cLSTM identifies the $k^\textrm{th}$ time series modality to be \textit{Granger non-casual} to the $j^\textrm{th}$ time-series modality. Otherwise, it is Granger-causal. In Equation~\ref{eq: GC_equation}, $W^{(j)}$ denotes the weights of the LSTM $L^{(j)}$ and $W_{:k}^{(j)}$ denotes the $k^\textrm{th}$ column of $W^{(j)}$.
\begin{table*}[t]
    \centering
    \caption{\small{\textbf{Dataset Information: }} We summarise details of the three datasets we used and evaluated~\modelname~on. We provide details about the dataset splits, features used and the evaluation metrics. We also mention emotion labels~(C:continuous and D:discrete) available and used for training.  }
    \label{tab:datasets}
    \resizebox{.8\textwidth}{!}{%
    \begin{tabular}{cccccccccc}
    \toprule
    && \multicolumn{6}{c}{Affective Features} & Labels & Evaluation \\
    \cmidrule{3-8}
   Dataset&Train/Val/Test &Facial& Audio & Textual & VA & Scene & Situation & (C/D) & Metrics \\
    \midrule

    SEND~\cite{ong2019computational} & 144/39/41 videos &\checkmark & \checkmark & \checkmark & $\times$& $\times$ &$\times$ & Valence~(C) & CCC\\
    % \midrule
    MovieGraphs~\cite{moviegraphs}  & 33/7/10 movies &\checkmark & \checkmark & \checkmark & \checkmark& \checkmark &\checkmark & $26$ classes~(D) & Accuracy\\
    % \midrule
    LIRIS-ACCEDE~\cite{baveye2015liris} & 44/10/12 movies &\checkmark & \checkmark & $\times$ & \checkmark& \checkmark &$\times$ & Valence/Arousal~(C) & MSE\\
    \bottomrule
    \end{tabular}
    }
    \vspace{-10pt}
\end{table*}

%%%%%%%%%%%%%%%%%%%%%%%%%%%%%%%%%%%%%%%%%
\section{\modelname: Our Approach}
\label{sec:approach}
We first give an overview of our method and some notations used in Section~\ref{subsec:overview}. This is followed by a description of each of the individual components of \modelname~in Section~\ref{subsec:rep},~\ref{subsec:co-att},~\ref{subsec:encoder}, and \ref{subsec:decoder}. 

%%%%%%%%%%%%%%%%%%%%%%%%%%%%%%%%%%%%%%%%%%%%%%%%%%
\subsection{Overview}
\label{subsec:overview}
We present an overview of our time-series emotion prediction model for multimedia content, \modelname~in Figure~\ref{fig:network}. Our approach draws on the theory of Emotional Causality to infer the emotional state that is conveyed in the multimedia content. The theory of Emotion Causality~\cite{films-causal} consists of the following main events:
\begin{enumerate}
    \item Identifying Outer Event: This stage refers to a stimulus that is contained in the multimedia content that causes an emotion in the consumer of that multimedia content. Such emotion-causing stimuli in multimedia commonly include affective cues such as facial expressions and speech of the actors, but often also include the visual aesthetic features as well as the context of the multimedia content.
    
    We work along these lines and extract these affective cues and visual aesthetic features from the multimedia content and build an affective-rich representation, described further in Section~\ref{subsec:rep}.
    \item Perceiving the Emotional State: This event refers to the formation of an emotional state in the consumer of the multimedia content upon receiving the stimuli contained in the latter. We develop a novel co-attention-based deep neural network that predicts a perceived emotion conveyed through multimedia.
\end{enumerate}

Lastly, the theory discussed thus far assumes a notion of causality between the two events. That is, the ``outer event'' \textit{causes} the ``perceived emotional state''. In order to computationally model this causality, we investigate the causality between the affective cues using Granger causality~\cite{gc}. In the following sections, we describe the computational details of the two events of Emotion Causality.

%%%%%%%%%%%%%%%%%%%%%%%%%%%%%%%%%%%%%%%%%%%%%%%%%%%%
\subsection{Building the Affective-Rich Representation}
\label{subsec:rep}
Our goal here is to build a representation of features that are capable of inciting a perceived emotional state in the audience. We extract features that can contribute to an affective-rich representation from the input $\mathcal{C}^{1:T}$ clips of the multimedia content. For each clip, we extract at most $p$ feature representations, including but not limited to-- the facial action units of the actors involved, the audio features, embeddings of the text/transcript, scene descriptions, action or situation descriptions and visual aesthetics of the clips,
\begin{equation}
    f_i^{1:T}  = \mathcal{F}_i(\mathcal{C}^{1:T})
\end{equation}
\noindent where $f_i^{1:T} \in \mathbb{R}^{p \times \left \vert \textrm{feature size} \right \vert_i \times T}$ for $i = 1,2, \ldots, p$ is the $i^\textrm{th}$ feature representation obtained using the extractor, $\mathcal{F}_i$. We describe the feature extractor $\mathcal{F}$ in Section~\ref{subsec:feature-extraction}. \modelname~can work with any subset of the $6$ features representations mentioned. This is also shown in our results. 
%%%%%%%%%%%%%%%%%%%%%%%%%%%%%%%%%%%%%%%%%%%%%%%%%%%%
\subsection{Perceiving the Emotional State}
We use the features, $f_1^{1:T}, f_2^{1:T} \ldots f_p^{1:T}$, generated from the first event to predict perceived emotions in the consumer of the multimedia content. Our approach consists of a deep neural network that uses co-attention to learn and, ultimately, to be able to focus on the useful elements of the features at different time instances. The key intuition here is that the relevance of features varies during the length of the multimedia. For example, the scene-setting and the visual aesthetics may stand out towards the beginning of a movie, as the viewers are not acquainted with the actors, rather are trying to build the scene in their minds. But later, the facial expressions and the speech of the actors may develop a stronger presence as the viewers get to know the actors and their stories. Co-attention helps capture the time-varying nature of the pairwise interdependent affective-rich features, implicitly handling transitivity amongst related modalities.

We begin by simultaneously encoding the feature representation (Eqn.~\ref{eq: encoder}) using recurrent neural network architectures (cLSTMs in our case) and computing the co-attention between pairs of features (Eqn.~\ref{eq: alpha_k}). The results from these operations are convolved to obtain a final context vector (Eqn.~\ref{eq: decoder}). The final perceived emotion label is computed by passing the context vector through an LSTM decoder, followed by a combination of linear and non-linear matrix multiplication operations (Eqn.~\ref{eq: final_output}).

\paragraph{cLSTM Encoder:}
\label{subsec:encoder}
We encode the $p$ time-series features using a cLSTM. As described in Section~\ref{subsec:gc}, we first compute $X_{1:T}$ using Equation~\ref{eq: GC_representation}. More formally, the transformation, $\Pi(f_i^{1:T})$ is given by, 

\[\Pi(\cdot) = \text{softmax}({\phi(\cdot)}) \]

\noindent Then, the transformed feature representation of each time-series is computed as,

\begin{equation}
    % \begin{split}
    x_i^{1:T}  = \Pi(f_i^{1:T}),
    % & \vdots\\
    % h_n & = \text{softmax}(\phi(\tanh{\phi(f_n)}))\\
    % h &= h_1 \oplus h_2 \oplus \dots \oplus h_n
    % \end{split}
\end{equation}

\noindent where $\phi$ is a linear operation with suitable weights. We can then obtain $X^{1:T}$ by row-wise stacking $x_i^{1:T}$ as follows,

\[    X^{1:T} = x_1^{1:T} \oslash x_2^{1:T} \oslash \dots \oslash x_p^{1:T},
\]

\noindent where $\oslash$ is a row-wise stacking operator. The stacked inputs are encoded using the cLSTM encoder as defined in Eq.~\ref{eq: GC_equation}. 
% \begin{align}
    \begin{equation}
    h_\textrm{enc}  = \text{cLSTM}(X^{1:T})
    \end{equation}
    \label{eq: encoder}
% \end{align}
%%%%%%%%%%%%%%%%%%%%%%%%%%%%%%%%%%%%%%%%%%%%%%%%%%%%
\vspace{-15pt}
\paragraph{Co-attention Scores:}
\label{subsec:co-att}
We learn the interactions between the different features $f_1^{1:T}, f_2^{1:T} \ldots f_p^{1:T}$ by aligning and combining modalities pairwise using Eq.~\ref{eq:co-attention-mechanism}. We obtain $m$ values of $\alpha_k$, where $k = 1,2, \ldots, m$ and $m=$ $p\choose{2}$, corresponding to each pairwise co-attention operation.

% \begin{align}
%     \begin{split}
%     \alpha_1 & = \text{Co-attention}(g_1(f_1), g_2(f_2))\\
%     & \vdots\\
%     \alpha_m & = \text{Co-attention}(g_{n-1}(f_{n-1}), \phi(f_n))\\
%     \alpha &= \alpha_1 \oplus \alpha_2 \oplus \dots \oplus \alpha_m
%     \end{split}
% \end{align}

\begin{equation}
        \alpha_k = \text{Co-attention}(\phi(f_{k_1^{1:T}}), \phi(f_{k_2}^{1:T}))
        \label{eq: alpha_k}
\end{equation}
where $k_1, k_2$ are indices denoting one of the $p$ feature vectors, $\phi(.)$ are linear layer operators with appropriate weights and $\oplus$ is the concatenation operator. We obtain a final $\alpha$ as,
\[\alpha = \alpha_1 \oplus \alpha_2 \oplus \dots \oplus \alpha_m\]
%%%%%%%%%%%%%%%%%%%%%%%%%%%%%%%%%%%%%%%%%%%%%%%%%%%%
\paragraph{Decoder:}
\label{subsec:decoder}
Finally, once we have computed $h_{enc}$ and $\alpha$, we can obtain the `context vector'$d$, by convolving the attention weights~($\alpha$) with the encoded feature representation, $h_{enc}$,  
\begin{equation}
    d = h_\textrm{enc} \otimes \alpha 
    \label{eq: decoder}
\end{equation}
The idea is to retain more of the corresponding values of $h_{enc}$ where the attention weights $\alpha$ are high and less information from the parts where the attention weights are low. 
Finally, the decoder uses the `context vector', $d$ and the past emotion values $y'$, concatenated together. 
We simply concatenate these two vectors and feed the merged vector to the decoder. This returns $\hat{y}$ the predicted labels. 
\begin{align}
    h_\textrm{dec} &= \text{LSTM} (d \oplus y')\\
    \hat{y} &= \phi(\text{ReLU}(\phi( h_\textrm{dec})))
    \label{eq: final_output}
\end{align}

\noindent\textbf{Vector Auto Regressive (VAR) training of Shared cLSTM Encoder: }The cLSTM encoder is shared in a multitask learning fashion to regress future values of input multimodal time-series data $\{ f_1^{1:T},f_2^{1:T},\dots f_p^{1:T}\}$ through vector autoregressive training \cite{tank2018neural} as shown in Equation \ref{eq: GC_equation}. The VAR training can viewed as a secondary task to the primary emotion prediction task, involving shared encoder layers. The group lasso penalty applied to the columns of the $W^{(j)}_{:k}$ matrix forces the cLSTM to predict the future values of $k^{th}$ modality  without relying on the past values of $j^{th}$ modality. The ridge regularization penalty ($r$), parameter for non-smooth regularization ($l$) and the learning rate of VAR training determine the sparsity of the Granger Causal relationship matrix \cite{tank2018neural} to mitigate the problem of multicollinearity amongst the multivariate time series input.
\section{Implementation Details}
\label{sec:implementation}
We give an overview of the datasets~(SENDv1, LIRIS-ACCEDE, and MovieGraphs) used for evaluating \modelname~in Section~\ref{subsec:Datasets}. In Section~\ref{subsec:feature-extraction}, we mention the features extracted for training. Finally, to aid reproducibility of our work, we list the training hyperparameters in Section~\ref{subsec:hyperparameters}.
%%%%%%%%%%%%%%%%%%%%%%%%%%%%%%%%%%%%%%%%%%%%%%%%%%
%%%%%%%%%%%%%%%%%%%%%%%%%%%%%%%%%%%%%%%%%%%%%%%%%%
\subsection{Datasets}
\label{subsec:Datasets}
Here we discuss details of the three datasets we used to evaluate and benchmark \modelname. For further readability, we have summarized these details in Table~\ref{tab:datasets}.\\
%%%%%%%%%%%%%%%%%%%%%%%%%%%%%%%%%%%%%%%
\noindent \textbf{SENDv1 Dataset: }The dataset consists of video clips of people recounting important and emotional life stories unscripted. The videos have been recorded in a face-centered setting with no background. Valence ratings collected are provided for every $0.5$ seconds of the video. \\
\ul{Evaluation Metrics: }Most previous works in this dataset have reported the Concordance Correlation Coefficient~(CCC)~\cite{lawrence1989concordance} for validation and test splits along with the standard deviation. The CCC captures the expected discrepancy between the two vectors, compared to the expected discrepancy if the two vectors were uncorrelated and can be calculated as follows:
\begin{equation*}
    \text{CCC}(Y, \hat{Y}) = \frac{2 \text{ Corr}(Y, \hat{Y}) \sigma_Y \sigma_{\hat{Y}}}{\sigma_Y^2 \sigma_{\hat{Y}}^2 + (\mu_y - \mu_{\hat{Y}})^2}
\end{equation*}
where $\text{ Corr}(Y, \hat{Y})$ is the Pearson correlation between the groundtruth~($Y$) and predicted valence values~($\hat{Y}$) for $T$ clips/timestamps of a video, and the $\mu$ and $\sigma$ are the mean and standard deviation predictions.\\
%%%%%%%%%%%%%%%%%%%%%%%%%%%%%%%%%%%%%%%
\noindent \textbf{LIRIS-ACCEDE Dataset: }The dataset contains videos from a set of $160$ professionally made and amateur movies. The movies are in various languages including English, Italian, Spanish, and French.\\
\ul{Evaluation Metrics: }Consistent with prior methods we report the Mean Squared Error~(MSE) for the test splits. Given predicted valence value, $\hat{y}$ and true valence value $y$ for $T$ clips of a movie, we compute the MSE as follows:
\begin{equation*}
    \text{MSE} = \frac{1}{T} \sum_{i=1}^{N} \left ( y_i - \hat{y_i} \right )^2
\end{equation*}
%%%%%%%%%%%%%%%%%%%%%%%%%%%%%%%%%%%%%%%%%%%%%%
\noindent \textbf{MovieGraphs Dataset: }This dataset provides detailed graph-based annotations of social situations depicted in movie clips for $51$ popular movies. Each graph consists of several types of nodes to capture the emotional and physical attributes of actors, their relationships, and the interactions between them. The dataset was collected and manually annotated using crowd-sourcing methods. We process all clips in every movie and extract the available `emotional' attributes and group them in $26$ discrete emotion labels as described by Kosti et al.~\cite{Kosti_2017_CVPR_Workshops}. Details on creating processed discrete $26$ labels are listed in Section~\ref{subsec:mg_labels}.\\
\ul{Evaluation Metrics: }Because the labels in MovieGraphs are discrete, detailing $26$ emotion classes, we report and compare against the Top-1 accuracy.
%%%%%%%%%%%%%%%%%%%%%%%%%%%%%%%%%%%%%%%%%%%%%%%%%%
%%%%%%%%%%%%%%%%%%%%%%%%%%%%%%%%%%%%%%%%%%%%%%%%%%
\subsection{Feature Extraction}
\label{subsec:feature-extraction}
To build up an affective rich representation of the multimedia content, we work with a total of $6$ features: facial, audio, textual, visual aesthetics, scene, and situation descriptors. We summarize the features available for training \modelname~in Table~\ref{tab:datasets}. An in-depth discussion of these feature extraction steps specific to the datasets has been added in Section~\ref{subsec:detail-features}. 
%%%%%%%%%%%%%%%%%%%%%%%%%%%%%%%%%%%%%%%%%%%%%%%%%%
%%%%%%%%%%%%%%%%%%%%%%%%%%%%%%%%%%%%%%%%%%%%%%%%%%
\subsection{Training Hyperparameters}
\label{subsec:hyperparameters}
All our results were generated on an NVIDIA GeForce GTX1080 Ti GPU. Hyper-parameters for our model were tuned on the validation set to find the best configurations.We used RMSprop for optimizing our models with a batch size of $1$. We experimented with the range of our model's hyperparameters such as: number of hidden layers $(1, 2, 3)$, size of hidden layers (cLSTM, LSTM Decoder, Dense), dropout $\{0.2, 0.3, 0.4, 0.5.0.6\}$, hidden dimension $\{64, 128, 256, 512\}$, and embedding dimension $\{64, 128, 256, 512\}$. The ridge penalty ($r$) and non-smooth regularization parameter ($l$) of VAR training of the cLSTM was kept constant at $1e^{-4}$ and $0.001$, respectively.The learning rate of both the tasks - emotion prediction and VAR were in this range - $\{1e^{-5}, 1e^{-4}, 1e^{-3}, 1e^{-2}\}$. More specific details on model specific training hyperparameters are added in Section~\ref{subsec:detail-hyperparameters}. 

%%%%%%%%% RESULTS %%%%%%%%%
\section{Results}
\label{sec:results}
% \input{tables/datasets}
% \input{tables/SEND_Eval}
% \input{tables/MG_Eval}
% \input{tables/ME2018_Eval}
%%%%%%%%%%%%%%%%%%%%%%%%%%%%%%%%%%%%%%%%%%%%%%%%%%%
We first discuss our quantitative results in Section~\ref{subsec:quantitative}, where we compare the performance of our method with SOTA methods on the three datasets. We then go over the ablation experiments performed in Section~\ref{subsec:ablation}. Finally, in Section~\ref{subsec:qualitative}, we present some qualitative results for \modelname.
\subsection{Quantitative Results}
\label{subsec:quantitative}
We compare \modelname~with SOTA methods on the three datasets. \\
\noindent \textbf{SENDv1 Dataset: } For this dataset we summarise the results in Table~\ref{tab:send_eval}. We provide CCC score for every combination of the three features/modalities used and also the combined results. Two prior works~\cite{ong2019computational, 8913483} have experimented with various network architectures and ideas and reported CCC values. We list and compare our performance from their best-performing methods in Table~\ref{tab:send_eval}.\\
\noindent \textbf{MovieGraphs Dataset: } The dataset has not been previously tested for any similar task. We trained a recently proposed SOTA method, EmotionNet~\cite{wei2020learning}, for affective analysis of web images on MovieGraphs dataset and compared our performance with this approach. We summarize this result in Table~\ref{tab:moviegraphs_eval}.\\
\noindent \textbf{LIRIS-ACCEDE Dataset: } For this dataset we summarise the results in Table~\ref{tab:mediaeval2018_eval}. To be consistent with prior methods, we report Means Squared Error~(MSE) and Pearsons Correlation Coefficient~(PCC) for our method. We compare against $7$ existing SOTA methods evaluated on the same dataset. Some of these listed methods were a part of the MediaEval2018 Challenge.
%%%%%%%%%%%%%%%%%%%%%%%%%%%%%%%%%%%%%%%%%%%%%%%%%%
\subsection{Ablation Experiments}
\label{subsec:ablation}
We perform a small ablation study to analyze the importance of each of the two components we incorporate for modeling the temporal causality in \modelname. We report the performance without co-attention between the features and also without Granger causality. These results are summarised in Table~\ref{tab:ablation}. As reported, we see a performance improvement of about $4-5$\% across all datasets and another $2-3$\% with the addition of Granger causality. 
%%%%%%%%%%%%%%%%%%%%%%%%%%%%%%%%%%%%%%%%%%%%%%%%%%%
%%%%%%%%%%%%%%%%%%%%%%%%%%%%%%%%%%%%%%%%%%%%%%%%%%%
\begin{figure*}[t]
    \centering
    \includegraphics[width = .8\linewidth]{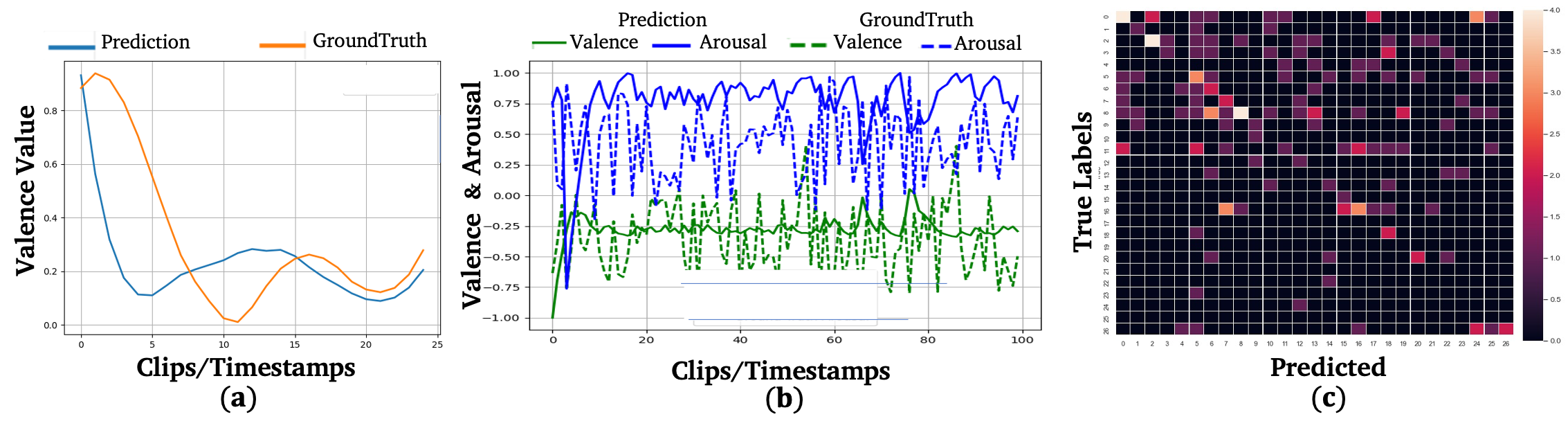}
    \caption{\small{\textbf{Qualitative Plots: }\textit{(a)} We show the valence values learned for a single sample on the test set in the SENDv1 Dataset, and in \textit{(b)}, we show both valence and arousal values from the LIRIS-ACCEDE Dataset. \textit{(c)} We show the confusion matrix for the MovieGraphs dataset. Brighter colors on the diagonal indicate more number of correct classifications. }}
    \label{fig:qual}
    \vspace{-11pt}
\end{figure*}
%%%%%%%%%%%%%%%%%%%%%%%%%%%%%%%%%%%%%%%%%%%%%%%%%%%
%%%%%%%%%%%%%%%%%%%%%%%%%%%%%%%%%%%%%%%%%%%%%%%%%%%
%%%%%%%%%%%%%%%%%%%%%%%%%%%%%%%%%%%%%%%%%%%%%%%%%%%
%%%%%%%%%%%%%%%%%%%%%%%%%%%%%%%%%%%%%%%%%%%%%%%%%%%
\begin{figure}[t]
    \centering
    \includegraphics[width = .8\columnwidth]{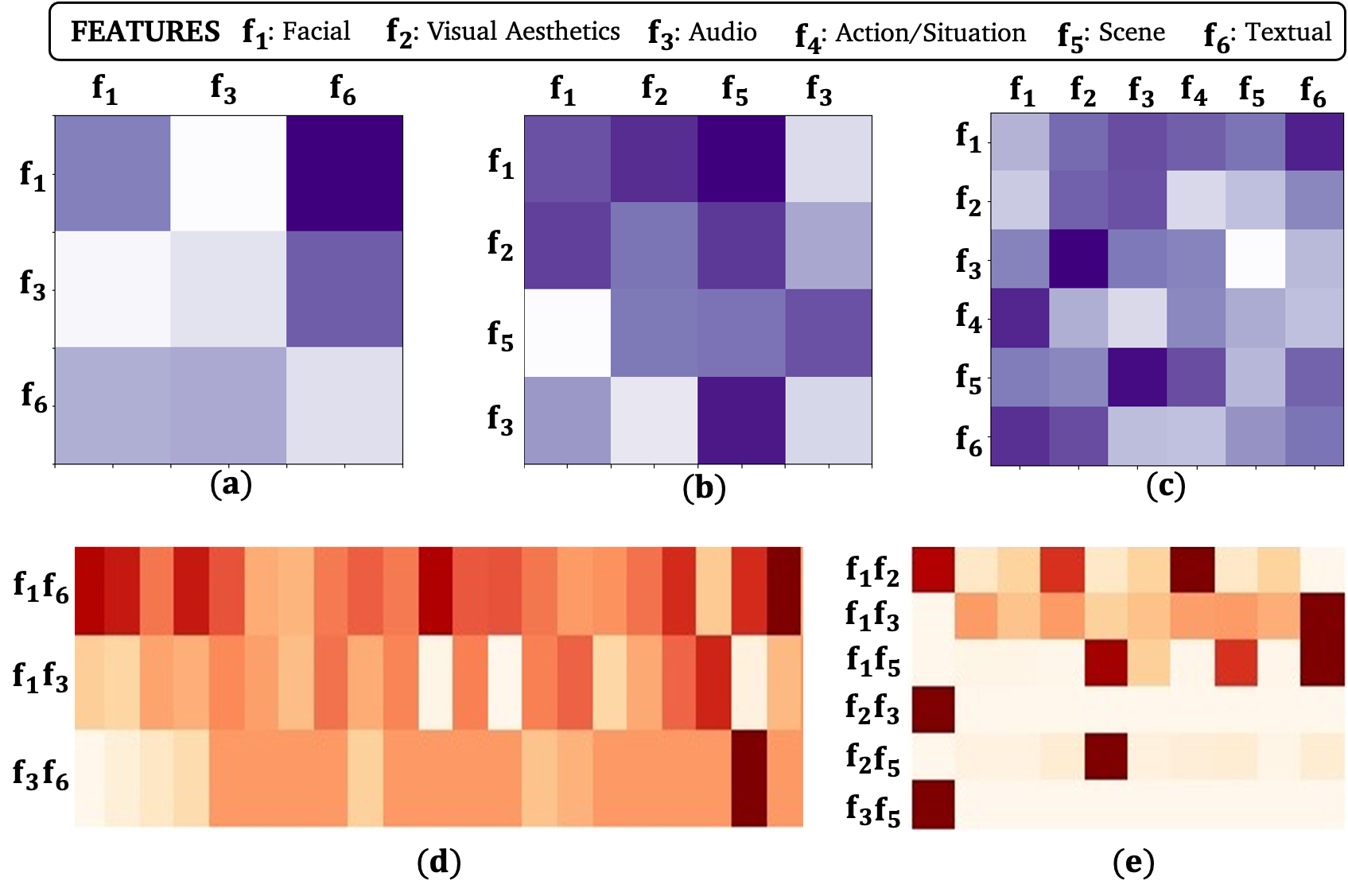}
    \caption{\small{\textbf{GC Matrices and Co-attention Weights:} \textit{(Top)} We show the GC matrices for all three datasets~\cite{ong2019modeling, baveye2015liris, moviegraphs}.
    % We also list the $f_1 \dots f_6$ features and respective features used for each dataset. 
    Darker shades indicate existence of a Granger causality between corresponding features.
    % For instance, in (a) $f_6$ \textit{Granger causes $f_1$} whereas $f_3$ \textit{Granger non-causes $f_1$}.
 \textit{(Bottom)} We plot normalized co-attention weights for one sample each from the (d) SENDv1 and (e) LIRIS-ACCEDE datasets with time-stamps on the $x$-axis and pairs of features on the $y$-axis. Darker shades imply a higher co-attention between the pair of the features and more relevance to predictions of the current and the following clips. }}
    \label{fig:gc-attn}
    \vspace{-12pt}
\end{figure}
%%%%%%%%%%%%%%%%%%%%%%%%%%%%%%%%%%%%%%%%%%%%%%%%%%%
%%%%%%%%%%%%%%%%%%%%%%%%%%%%%%%%%%%%%%%%%%%%%%%%%%%
\begin{table}[t]
    \centering
    \caption{\small{\textbf{Evaluation on SENDv1 Dataset: } for time-series emotion perception of \modelname. We report CCC values. }}
    \label{tab:send_eval}
    \resizebox{\columnwidth}{!}{%
    \begin{tabular}{rcccccccc}
    \toprule
    \textbf{Method}  & \multicolumn{8}{c}{\textbf{Modalities}}  \\
    \cmidrule{2-9}
    &  \multicolumn{2}{c}{\textbf{FA}} &  \multicolumn{2}{c}{\textbf{AT}} &  \multicolumn{2}{c}{\textbf{FT}} &  \multicolumn{2}{c}{\textbf{FAT}} \\
    \cmidrule{2-9}
    &  Val & Test & Val & Test & Val & Test & Val & Test \\
    \midrule
    LSTM~\cite{ong2019modeling}  & $.100$ & $.160$  & $.800$ & $.090$ & $.280$ & $.400$  & $.140$ & $.150$\\
    VRNN~\cite{ong2019modeling}  & $.140$ & $.170$  & $.320$ & $.350$  & $.240$ & $.300$ & $.170$ & $.240$ \\
    SFT~\cite{8913483}  & $.150$ & $.160$  & $.080$ & $.080$  & $.320$ & $.350$ & $.120$ & $.140$  \\
    MFT~\cite{8913483}  & $.060$ & $.080$  & $.360$ & $.330$  & $.400$ & $.360$  & $.420$ & $.440$ \\
    B3-MFN~\cite{8913483} & $0.220$ & $.090$  & $.370$ & $.330$  & $.330$ & $.310$  & $.340$ & $.280$  \\
    Human~\cite{ong2019modeling}  & - & - & - & - & - & - & $.470$ & $.500$\\
    \rowcolor{blue!20} \textbf{Ours}  & \textbf{$.557$} &\textbf{ $.582$} & \textbf{$.592$}& \textbf{$.601$} & \textbf{$.556$} &\textbf{ $.567$} & \textbf{$.599$} & \textbf{$.597$} \\
    \bottomrule
    \end{tabular}
    }
\end{table}

\begin{table}[t]
    \centering
    \caption{\small{\textbf{Evaluation on MovieGraphs: }}for time-series emotion perception of \modelname for Affective Analysis of Multimedia. We report top-1 accuracy for comparisons.}
    \label{tab:moviegraphs_eval}
    \resizebox{.8\columnwidth}{!}{%
    \begin{tabular}{ccc}
    \toprule
    Method & Validation &  Test \\
     & (Top-1 Acc) & (Top-1 Acc)\\
    \midrule
EmotionNet~\cite{wei2020learning} & $35.60$ & $27.90$\\
    \rowcolor{blue!20} \textbf{Ours} & \textbf{$39.88$} & \textbf{$30.58$}\\
    \bottomrule
    \end{tabular}
    }
    \vspace{-10pt}
\end{table}
% \begin{table}[th]
%     \centering
%     \caption{\small{\textbf{Evaluation on MediaEval2018 Dataset: } for time-series emotion perception of \modelname for Affective Analysis of Multimedia. We report MSE and PCC for comparisons.}}
%     \label{tab:mediaeval2018_eval}
%     % \resizebox{\columnwidth}{!}{%
%     \begin{tabular}{ccccc}
%     \toprule
%     & \multicolumn{2}{c}{\textbf{Valence}}  & \multicolumn{2}{c}{\textbf{Arousal}}  \\
%     \cmidrule{2-5}
%     Method& MSE&PCC &MSE&PCC\\
%     \midrule
%  CERTH-ITI~\cite{batziou2018visual} & $0.117$&$0.098$ & $0.138$ & $0.054$ \\
%  THUHCSI~\cite{jin2017thuhcsi} &$0.092$&$0.305$ &  $0.140$& $0.087$\\
%  Quan et al.~\cite{quan2018frame} &$0.115$&$0.146$&  $0.171$& $0.091$\\
%  Yi et al.~\cite{yi2018cnn} &$0.090$&$0.301$&  $0.136$& $0.175$\\
%  GLA~\cite{sun2019gla} &$0.084$&$0.278$&  $0.133$& $0.351$\\
%  Ko et al.~\cite{ko2018towards}  &$0.102$&$0.114$&  $0.149$& $0.083$\\
%  Zhao et al~\cite{zhao2019video}&$0.071$&$0.444$&  $0.137$& $0.419$\\
%   \rowcolor{blue!20} \textbf{Ours}   & 0.068 & & 0.128& \\
%     \bottomrule
%     \end{tabular}
%     % }
%     \vspace{-10pt}
% \end{table}

\begin{table}[t]
    \centering
    \caption{\small{\textbf{Evaluation on LIRIS-ACCEDE~(MediaEval2018 Dataset): } for time-series emotion perception of \modelname for Affective Analysis of Multimedia. We report MSE for comparisons~(lower the better).}}
    \label{tab:mediaeval2018_eval}
    \resizebox{.6\columnwidth}{!}{%
    \begin{tabular}{rcc}
    \toprule
    & {\textbf{Valence}}  & {\textbf{Arousal}}  \\
    Method& (MSE) &(MSE)\\
    \midrule
 CERTH-ITI~\cite{batziou2018visual} & $0.117$&  $0.138$  \\
 THUHCSI~\cite{jin2017thuhcsi} &$0.092$&  $0.140$\\
 Quan et al.~\cite{quan2018frame} &$0.115$&  $0.171$\\
 Yi et al.~\cite{yi2018cnn} &$0.090$&  $0.136$\\
 GLA~\cite{sun2019gla} &$0.084$&  $0.133$\\
 Ko et al.~\cite{ko2018towards}  &$0.102$&  $0.149$\\
 Zhao et al~\cite{zhao2019video}&$0.071$&  $0.137$\\
  \rowcolor{blue!20} \textbf{Ours}   & \textbf{$0.068$} & \textbf{$0.128$}\\
    \bottomrule
    \end{tabular}
    }
    \vspace{-10pt}
\end{table}

\begin{table}[t]
    \centering
    \caption{\small{\textbf{Ablation Studies: } We perform ablation experiments to understand the importance of co-attention and Granger Causality for modeling temporal causality.}}
    \label{tab:ablation}
    \resizebox{.8\columnwidth}{!}{%
    \begin{tabular}{ccccc}
    \toprule
    \textbf{Experiment} & \textbf{SENDv1} & \textbf{MG} &  \multicolumn{2}{c}{\textbf{LIRIS-ACCEDE}}\\
    &  &  & Valence & Arousal\\
    & (CCC) & (Acc) & (MSE) & Arousal(MSE)\\
    \midrule
    \modelname~w/o (co-attn.\& GC) &$.570$ & $28.290$ & $0.122$ &  $0.143$\\
    \modelname~w/o GC & $.585$ & $29.450$ & $0.095$ & $0.135$ \\
    \rowcolor{blue!20}\textbf{\modelname} & $.597$ & $30.580$ & $0.068$ & $0.128$ \\
    \bottomrule
    \end{tabular}
    }
    \vspace{-10pt}
    \end{table}

\subsection{Qualitative Results}
\label{subsec:qualitative}
\noindent\textbf{Time-Series Emotion Predictions: }
We present qualitative results for \modelname~in Figures~\ref{fig:qual} and~\ref{fig:gc-attn}. In Figure~\ref{fig:qual}(a), we show the valence values~(range $\in \left [ 0,1 \right ]$) learned for a single sample from the test set for SENDv1 Dataset. In Figure~\ref{fig:qual}(b), we show the predicted valence and arousal values for a sample from LIRIS-ACCEDE Dataset~(range $\in \left [ -1,1 \right ]$) along with groundtruth labels. In Figure~\ref{fig:qual}(c), we plot the $27 \times 27$ confusion matrix for all the test points in MovieGraphs dataset. The horizontal axis represents the predicted class labels while the vertical axis represents the true labels. The discrete classes are in alphabetical order with 27th class as `None'~(Section \ref{subsec:mg_labels}). Brighter colors on the diagonal indicate more correct classifications. \\
\noindent \textbf{Interpreting GC matrix: }We visualize the Granger causality plots in Figure~\ref{fig:gc-attn} (\textit{(a), (b) and (c)}) for all the three datasets. The dimensions of the GC matrix is $\left | \text{features} \times \text{features} \right |$. Hence, it is $\left | 3 \times3 \right |$ for SENDv1 dataset, $\left | 4 \times 4 \right |$ for LIRIS-ACCEDE dataset and $\left |6 \times 6 \right |$ for Moviegraphs dataset. To interpret the GC matrix, we read the features appearing on the horizontal axis and query whether they Granger-cause (or non-cause) the features on the vertical axis. Darker shades of purple indicate causality while lighter shades indicate non-causality. So for example, in (a) $f_3$ Granger-non-causes $f_1$ for SENDv1 dataset, and in (b), $f_5$ Granger-causes $f_1$ with a higher certainty.\\
\noindent \textbf{Causality Qualitative Analysis: }Here we further analyse co-attention and GC qualitatively. We present the analysis and reasoning of GC and co-attention for one such video from SENDv1 dataset~(ID123vid1, timestamp 1:52 in supplementary video), for which the corresponding GC matrix and co-attention weights are shown in Figures~\ref{fig:gc-attn}(a) and (d). \\
\noindent \textbf{Interpreting Co-attention Weights: }We visualize the co-attention weights for one sample each from SENDv1 and LIRIS-ACCEDE in Figure~\ref{fig:gc-attn} (\textit{(d), (e)}). We plot the clips/timestamps on the x-axis and pairs of all the features on the y-axis. For every clip, darker shades of red indicate a higher co-attention weight between a pair of modalities.
%
% the weights indicate the relevance they had towards the prediction of the emotion label for the current and the following clip/timestamp. Darker shades of red indicate a higher co-attention weight between the pair of modalities. \\
\begin{enumerate}[noitemsep, leftmargin=*]
    \item \textbf{Text GC Face:} In  Figure~\ref{fig:gc-attn}a, the text modality strongly Granger-causes the facial modality because at 2:20, when the subject says, ``... he had a sailor's mouth ...'', the subject then moments later motions to his face and changes his facial expression to mimic a sailor face.
    \item \textbf{Text GC Speech:} In  Figure~\ref{fig:gc-attn}a, the text Granger-causes speech which is most clearly evidenced when at 2:05, the subject says, ``... feel really lucky, really blessed ...'', then subject then audibly sighs with relief. 
    \item \textbf{Speech correlation with text:} In Figure~\ref{fig:gc-attn}d, we see a weak correlation between speech and text in the beginning because when the subject is remembering the good things about his grandfather, the text contains positive stories about the grandfather while the tone in his speech is still sad. However, the correlation between text and speech is stronger at the end of the video when the subject mentions the word, ``death", and then takes a long pause signifying great sadness and loss. 
\end{enumerate}
%%%%%%%%% CONCLUSION %%%%%%%%%
\section{Conclusion, Limitations and Future Work}
\label{sec:conclusion}
We present \modelname, a learning method for time-series emotion prediction for multimedia content. We use Emotional Causality via co-attention and Granger causality 
% to align our computational model with the psychology literature and also present two ideas to model temporal causality, co-attention, and Granger causality for
for modeling temporal dependencies in the input data and building an affective-rich representation to understand and perceive the scene. We evaluated our method on three benchmark datasets and achieved state-of-the-art results.

There are some limitations with our approach. We need to make the feature extraction process online. Furthermore, our approach does not currently work with scenes containing single actors
% . It will be interesting to use similar analysis on a per-person/actor basis in a video to automate analysis of debates and group discussions.
In the future, we would like to extend our approach to solve Step 3 of the Emotional Causality Theory to predict the physiological response of the viewer to the movie. We would also like to explore causality and correlation of modalities in single-actor scenes to automate analysis of debates and group discussions. Finally, we will incorporate the ideas presented in this paper with recommendation systems
\vspace{-10pt}

\section{Acknowledgements}
This work was supported in part by ARO Grants W911NF1910069, W911NF1910315, W911NF2110026  and Intel.

{\small
\bibliographystyle{ieee_fullname}
\bibliography{egbib}
}
\newpage
\appendix
\section{Training Details}
\label{appendix:1}
We expand and provide a more detailed look into the feature extraction for each of the datasets in Section~\ref{subsec:detail-features}. To ensure reproducability of our work, we provide extensive details into all training hyperparameters for all the datasets. 
\subsection{Feature Extraction}
\label{subsec:detail-features}
\subsubsection{SENDv1 Dataset}
We used the facial features, audio features and the text embeddings as input for SENDv1. We used the extracted features for the three modalities as explained by Ong et al.~\cite{ong2019modeling}. To summarize, for audio features they used openSMILE v$2.3.0$~\cite{eyben2013recent} to extract the extended GeMAPS (eGeMAPS) set of $88$ parameters for every $0.5$-second window.
For text features, they provide third-party commissioned professional transcripts for the videos. The transcript was then aligned~(every $5$ seconds) and a $300$-dimensional GloVe word embeddings~\cite{pennington2014glove} was used. For the facial features they provide $20$ action points~\cite{ekman1997face} extracted using the Emotient software by iMotions~\footnote{https://imotions.com/emotient/} for each frame~($30$ per second).\\ 
%%%%%%%%%%%%%%%%%%%%%%%%%%%%%%%%%%%%%%%%
\subsubsection{LIRIS-ACCEDE Dataset}
Like mentioned in Table~\ref{tab:datasets}, we used the facial features, audio features, scene descriptors and visual aesthetic features. While we used the already available features for audio and visual aesthetics, we extract the facial features and scene descriptors ourselves. The audio features provided were extracted using the openSMILE toolbox~\footnote{http://audeering.com/technology/opensmile/}, which compute a $1,582$ dimensional feature vector. For the visual aesthetics, the authors provide the following: Auto Color Correlogram, Color and Edge Directivity Descriptor, Color Layout, Edge Histogram, Fuzzy Color and Texture Histogram, Gabor, Joint descriptor joining CEDD and FCTH in one histogram, Scalable Color, Tamura, and Local Binary Patterns extracted using the LIRE~\footnote{ http://www.lire-project.net/} library. We extracted the face features ourselves using Bulat et al.~\cite{bulat2017far}. These result in $68$ action units with the 3D coordinates. For the scene descriptors we Xiao et al.'s~\cite{xiao2018unified} $4096$ dimensional intermediate representation.\\
% \usepackage{booktabs}
% \usepackage[normalem]{ulem}
\begin{table}[t]
\caption{\small{\textbf{MovieGraphs Labels Generation: } We list the attributes of Moviegraphs used for all clips for grouping them into $26$ discrete emotion labels. }}
\label{tab:mg_label_details}
\resizebox{\columnwidth}{!}{
\begin{tabular}{|l|l|l|}
\toprule
\textbf{Class Id} &\textbf{Emotion Labels} & \textbf{Attribute Labels available in MovieGraphs}                                                                                                                                                                  \\ \midrule
0 & Affection               & loving, friendly                                                                                                                                                                                                    \\ \midrule
1 & Anger                   & anger, furious, resentful, outraged, vengeful                                                                                                                                                                       \\ \midrule
2 & Annoyance               & \begin{tabular}[c]{@{}l@{}}annoy, frustrated, irritated, agitated, bitter, insensitive, \\ exasperated, displeased\end{tabular}                                                                                     \\ \midrule
3 & Anticipation            & optimistic, hopeful, imaginative, eager                                                                                                                                                                             \\ \midrule
4 & Aversion                & disgusted, horrified, hateful                                                                                                                                                                                       \\ \midrule
5 & Confident               & \begin{tabular}[c]{@{}l@{}}confident, proud, stubborn, defiant, independent, \\ convincing\end{tabular}                                                                                                             \\ \midrule
6 & Disapproval             & \begin{tabular}[c]{@{}l@{}}disapproving, hostile, unfriendly, mean, disrespectful, \\ mocking, condescending, cunning, manipulative, nasty, \\ deceitful, conceited, sleazy, greedy, rebellious, petty\end{tabular} \\ \midrule
7 & Disconnection           & \begin{tabular}[c]{@{}l@{}}indifferent, bored, distracted, distant, uninterested, \\ self-centered, lonely, cynical, restrained, unimpressed, \\ dismissive\end{tabular}                                            \\ \midrule
8 & Disquietment            & \begin{tabular}[c]{@{}l@{}}worried, nervous, tense, anxious, afraid, alarmed, \\ suspicious, uncomfortable, hesitant, reluctant, \\ insecure, stressed, unsatisfied, solemn, submissive\end{tabular}                \\ \midrule
9 & Doubt/Conf              & confused, skeptical, indecisive                                                                                                                                                                                     \\ \midrule
10 & Embarrassment           & embarrassed, ashamed, humiliated                                                                                                                                                                                    \\ \midrule
11 & Engagement              & \begin{tabular}[c]{@{}l@{}}curious, serious, intrigued, persistent, interested, \\ attentive, fascinated\end{tabular}                                                                                               \\ \midrule
12 & Esteem                  & respectful, grateful                                                                                                                                                                                                \\ \midrule
13 & Excitement              & \begin{tabular}[c]{@{}l@{}}excited, enthusiastic, energetic, playful, impatient, \\ panicky, impulsive, hasty\end{tabular}                                                                                          \\ \midrule
14 & Fatigue                 & tire, sleepy, drowsy                                                                                                                                                                                                \\ \midrule
15 & Fear                    & scared, fearful, timid, terrified                                                                                                                                                                                   \\ \midrule
16 & Happiness               & \begin{tabular}[c]{@{}l@{}}cheerful, delighted, happy, amused, laughing, thrilled, \\ smiling, pleased, overwhelmed, ecstatic, exuberant\end{tabular}                                                               \\ \midrule
17 & Pain                    & pain                                                                                                                                                                                                                \\ \midrule
18 & Peace                   & \begin{tabular}[c]{@{}l@{}}content, relieved, relaxed, calm, quiet, \\ satisfied, reserved, carefree\end{tabular}                                                                                                   \\ \midrule
19 & Pleasure                & \begin{tabular}[c]{@{}l@{}}funny, attracted, aroused, hedonistic, \\ pleasant, flattered, entertaining, mesmerized\end{tabular}                                                                                     \\ \midrule
20 & Sadness                 & \begin{tabular}[c]{@{}l@{}}sad, melancholy, upset, disappointed, discouraged, \\ grumpy, crying, regretful, grief-stricken, depressed, \\ heartbroken, remorseful, hopeless, pensive, miserable\end{tabular}        \\ \midrule
21 & Sensitivity             & apologetic, nostalgic                                                                                                                                                                                               \\ \midrule
22 & Suffering               & \begin{tabular}[c]{@{}l@{}}offended, hurt, insulted, ignorant, disturbed, \\ abusive, offensive\end{tabular}                                                                                                        \\ \midrule
23 & Surprise                & \begin{tabular}[c]{@{}l@{}}surprise, surprised, shocked, amazed, startled, \\ astonished, speechless, disbelieving, incredulous\end{tabular}                                                                        \\ \midrule
24 & Sympathy                & \begin{tabular}[c]{@{}l@{}}kind, compassionate, supportive, sympathetic, \\ encouraging, thoughtful, understanding, generous, \\ concerned, dependable, caring, forgiving, reassuring, gentle\end{tabular}          \\ \midrule
25 & Yearning                & \begin{tabular}[c]{@{}l@{}}jealous, determined, aggressive, desperate, focused, \\ dedicated, diligent\end{tabular}   \\ \midrule
26 & None                & \begin{tabular}[c]{@{}l@{}}-\end{tabular}   \\ \bottomrule
\end{tabular}}

\end{table}
\begin{table*}[]
\begin{tabular}{l|c|c|c}
\hline
\multicolumn{1}{l|}{}                  & \multicolumn{3}{c}{Dataset}                                                                       \\ \hline
\multicolumn{1}{c|}{Hyperparameters}   & \multicolumn{1}{c|}{SENDv1} & \multicolumn{1}{c|}{MovieGraphs} & \multicolumn{1}{c}{LIRIS-ACCEDE} \\ \hline
Dropout Ratio                           & 0.5                         & 0.5                              & 0.5                               \\
Optimizer                               & Adam                        & Adam                             & Adam                              \\
Embedding Dimension (Facial Expression) & 32                          & 204                              & 204                               \\
Embedding Dimension (Visual Aesthetics) & N/A                         & 41                               & 317                               \\
Embedding Dimension (Audio)             & 88                          & 300                              & 1584                              \\
Embedding Dimension (Action/Situation)  & N/A                         & 300                              & N/A                               \\
Embedding Dimension (Scene)             & N/A                         & 300                              & 4096                              \\
Embedding Dimension (Textual)           & 300                         & 300                              & N/A                               \\
Hidden Dimension (Linear Layers)        & 512                         & 1024                             & 512                               \\
Hidden Dimension (cLSTM Encoder)        & 512                         & 1024                             & 512                               \\
Hidden Dimension (LSTM Decoder)         & 512                         & 1024                             & 512                               \\
Number of hidden layers                 & 1                           & 1                                & 1                                 \\
Epochs                                  & 10                          & 10                               & 20                                \\
Batch Size                              & 1                           & 1                                & 1                                 \\
Learning Rate (Affect2MM model)         & 1e-4                        & 1e-4                             & 1e-4                              \\
Learning Rate (Multivariate VAR)        & 0.001                       & 0.001                            & 0.001                             \\
Activation Function of Linear layers    & LeakyReLU                   & LeakyReLU                        & LeakyReLU                         \\
Dimension of FCN Layers                 & {[}(512 x 4), (4 x 1){]}    & {[}(1024 x 4), (4 x 27){]}       & {[}(512 x 4), (4 x 2){]}     
\\ \hline
\end{tabular}
\caption{\textbf{Hyperparameters Details: }\label{hyperparameters}Training hyperparameters for SENDv1, MovieGraph and LIRIS-ACCEDE dataset.}
\end{table*}

%%%%%%%%%%%%%%%%%%%%%%%%%%%%%%%%%%%%%%%%
\subsubsection{MovieGraphs Dataset}
For MovieGraphs dataset as summarized in Table~\ref{tab:datasets}, we use all the features except the audio as the audios were not provided in the dataset. We now explain below how we retrieve the features. We extracted the face features ourselves using Bulat et al.~\cite{bulat2017far}. These result in $68$ action units with the 3D coordinates. For the transcript, we used the $300$-dimensional GloVe word embeddings~\cite{pennington2014glove} to obtain the feature representation. For visual aesthetic features, we extracted various features for color, edges, boxes and segments using Peng et al.~\cite{peng2018feast}. For the scene and situation descriptors we used the provided text in the dataset and used the $300$-dimensional GloVe word embeddings again to make them into feature representations. 
%%%%%%%%%%%%%%%%%%%%%%%%%%%%%%%%%%%%%%%%%%%%%%%%%%%%%%%%%%%%%%%%%%%%%%%%%%%%%%%%%%%%%%%%%%%%%%%%%%%%%%%%%%
\subsection{Training Hyperparameters}
\label{subsec:detail-hyperparameters}
Table \ref{hyperparameters} lists all the necessary dataset-specific training hyperparameters used in the proposed Affect2MM model.
%%%%%%%%%%%%%%%%%%%%%%%%%%%%%%%%%%%%%%%%
%%%%%%%%%%%%%%%%%%%%%%%%%%%%%%%%%%%%%%%%
\subsection{MovieGraphs Dataset Labels Generation}
\label{subsec:mg_labels}
We provide the attribute values we used as ``emotional keywords'' to group into the $26$ emotion labels that were then used for training MovieGraphs Dataset in table~\ref{tab:mg_label_details}.
%%%%%%%%%%%%%%%%%%%%%%%%%%%%%%%%%%%%%%%%
%%%%%%%%%%%%%%%%%%%%%%%%%%%%%%%%%%%%%%%%
\section{Codes}
To enable reproducability and further research in this domain we release codes for~\modelname~at \url{https://github.com/affect2mm/emotion-timeseries}.

\end{document}